\documentclass[10pt, journal]{IEEEtran}
\usepackage{cite}
\usepackage[tight,footnotesize]{subfigure}
\usepackage[cmex10]{amsmath}
\usepackage{amssymb}
\usepackage{mathtools}
\usepackage{multirow}
\usepackage{notoccite}
\usepackage{color}
\usepackage{epsfig}
\usepackage{latexsym}
\usepackage{subfigure}
\usepackage{pstricks}
\usepackage{pdfpages}
\usepackage{epstopdf}
\DeclareGraphicsExtensions{.eps}
\usepackage{algorithmic}
\usepackage{algorithm}
\usepackage{multirow}
\usepackage{graphicx}
\usepackage{xurl}
\usepackage{tabularx}
\usepackage{hyperref}
\usepackage{subcaption}

\begin{document}
	\title{Neural Precoding in Complex Projective Spaces}
	\author{\IEEEauthorblockN{
    Zaid Abdullah, M$\acute{\text{e}}$rouane Debbah,~\IEEEmembership{Fellow,~IEEE},\\ Symeon Chatzinotas,~\IEEEmembership{Fellow,~IEEE}, and Bj$\ddot{\text{o}}$rn Ottersten,~\IEEEmembership{Fellow,~IEEE}}
    \thanks{\par This work has been supported by the Smart Networks and Services
            Joint Undertaking (SNS JU) project TERRAMETA under the European Union’s Horizon Europe research and innovation programme
            under Grant Agreement No 101097101, including top-up funding by UK Research and Innovation (UKRI) under the UK government’s Horizon Europe funding guarantee. 
            \par Z. Abdullah, S. Chatzinotas, and B. Ottersten are with the Interdisciplinary Centre for Security, Reliability, and Trust (SnT) at the University of Luxembourg, Luxembourg. 
            \par M. Debbah is with the 6G Research Center, Khalifa University of Science and Technology, Abu Dhabi, UAE.}}
	
\maketitle
\begin{abstract}
Deep-learning (DL)-based precoding in multi-user multiple-input single-output (MU-MISO) systems involves training DL models to map features derived from channel coefficients to labels derived from precoding weights. Traditionally, complex-valued channel and precoder coefficients are parameterized using either their real and imaginary components or their amplitude and phase. However, precoding performance depends on magnitudes of inner products between channel and precoding vectors, which are invariant to global phase rotations. Conventional representations fail to exploit this symmetry, leading to inefficient learning and degraded generalization. To address this, we propose a DL framework based on complex projective space (CPS) parameterizations of both the wireless channel and the weighted minimum mean squared error (WMMSE) precoder vectors. By removing the global phase redundancies inherent in conventional representations, the proposed framework enables the DL model to learn geometry-aligned and physically distinct channel-precoder mappings. Two CPS parameterizations based on real-valued embeddings and complex hyperspherical coordinates are investigated and benchmarked against two baseline methods. Simulation results demonstrate substantial improvements in sum-rate performance and generalization, with negligible increase in model complexity.
\end{abstract}
\section{Introduction}\label{sec:intro}
\IEEEPARstart{S}{patial} multiplexing in modern wireless systems enables multiple users to be served simultaneously using shared frequency resources. To mitigate inter-user interference, the multi-antenna base station (BS) must process the signals before transmission. This is commonly known as \textit{precoding}, and it involves weighting the transmit signals with carefully designed coefficients, typically derived from the estimated channel state information (CSI). 

Various algorithms have been proposed for precoding design over the years. However, optimal solutions often necessitate non-linear processing with prohibitive computational complexity, rendering such methods impractical for real-time deployment. Consequently, practical systems often rely on sub-optimal, low-complexity linear precoders with closed-form solutions. More recently, a significant body of literature has emerged on leveraging deep learning (DL) to approximate near-optimal solutions while maintaining low computational overhead during the inference phase. In this approach, the DL model is trained to find a mapping between input features extracted from the known CSI and output labels derived from the precoding coefficients. 

In the following, we summarize the main contributions in the literature on conventional- and DL-based precoding, before introducing our proposed complex projective space (CPS) neural precoding framework.
\subsection{Previous Works}
\subsubsection{Conventional Precoding Schemes}
Various precoding objectives have been investigated in the literature. A common formulation minimizes the total transmit power at the BS while satisfying minimum quality-of-service constraints for all users (UEs). As shown in~\cite{SB-9931, SB-0105}, the solution for this power-minimization problem can be obtained using interior-point methods (IPM) for semidefinite optimization.

Another important objective is the maximization of a utility function, such as the achievable sum-rate, subject to a total power constraint. In contrast to the power-minimization problem, sum-rate maximization is considerably more challenging and has been proven to be NP-hard, despite having a simple solution structure~\cite{6832894}. A more flexible formulation considers weighted sum-rate (WSR) maximization to enable user prioritization. Although the WSR problem is also NP-hard, it is commonly addressed using the weighted minimum mean squared error (WMMSE) algorithm. By exploiting the equivalence between WSR maximization and WMMSE minimization, this iterative framework enables closed-form updates for each block variable and is guaranteed to converge to a stationary point. Nevertheless, the WMMSE algorithm remains computationally demanding due to repeated matrix inversions across multiple iterations.

To reduce complexity, closed-form linear precoders are often employed. Two widely adopted techniques are maximum ratio transmission (MRT) and zero-forcing (ZF)~\cite{yang2013performance}. However, these approaches exhibit inherent limitations. Specifically, MRT becomes interference-limited at high signal-to-noise ratio (SNR), whereas ZF suffers from power inefficiency at low SNR due to channel inversion. Moreover, ZF is sensitive to CSI errors and experiences performance degradation when the number of UEs becomes large~\cite{abdullah2022low}.
\subsubsection{DL-based Precoding Schemes}
The main motive behind designing DL-based precoding is to approximate the solutions of high-complexity, optimization-based methods (such as the WMMSE precoder or those relying on IPM) while ensuring affordable complexity during inference. 

The works in~\cite{vu2021machine, wang2025learning} investigated a DL-based joint design of antenna selection and precoding in co-located and distributed systems, respectively. In addition, DL models were designed in~\cite{elbir2019hybrid, liu2025deep} for hybrid analog-digital precoding with the objective of maximizing the achievable rate. In a similar research direction, the authors in~\cite{liu2022dl} designed DL-based hybrid precoding for maximizing the energy efficiency. 

To improve learning efficiency and by relying on the known optimal precoder structure, the works in~\cite{kim2020deep, zhang2022deep, shi2023robust, shi2021deep} proposed a DL approach where the model predicts key features for constructing the optimal precoders instead of predicting the full precoding matrix. 

Unlike the works in~\cite{vu2021machine, wang2025learning, elbir2019hybrid, liu2025deep, liu2022dl, kim2020deep, zhang2022deep, shi2023robust, shi2021deep} that mainly focused on algorithmic and/or network architecture aspects, our recent work~\cite{abdullah2025impact} provided key perspectives and insights on DL-based precoding design. Specifically, we investigated how different preprocessing methods of input features derived from the known CSI impact the underlying channel statistics, error performance, and model stability. Notably, our empirical analysis demonstrated that preprocessing and model complexity should be tackled jointly rather than treated as separate design factors. 
\subsection{Contributions}
In DL-based precoding, the input features of a DL model, such as a deep neural network (DNN), are derived from the CSI coefficients, while the labels are derived from the precoding weight coefficients. Existing works usually represent these complex-valued channel and precoding coefficients using (i) real and imaginary components, (ii) amplitude (absolute value) and phase, or (iii) a combination of the two~\cite{vu2021machine, wang2025learning, elbir2019hybrid, liu2025deep, liu2022dl, kim2020deep, zhang2022deep, shi2023robust, shi2021deep, abdullah2025impact}.

However, for multi-user multiple-input single-output (MU-MISO) systems, precoding performance depends on the magnitudes of inner products between complex-valued channel and precoding vectors, which are invariant to global phase rotations. As a result, conventional parameterizations require the DNN to implicitly learn this phase invariance, introducing redundancies and increasing the learning difficulty.

To address this, we propose a DL framework where channel and precoder vectors are mapped to an $n$-dimensional complex projective space, denoted as $\mathbb {CP}^n$. In this space, vectors differing only by a global phase rotation are equivalent. By deriving features and labels from these representative points on the manifold, our approach enables learning directly on the space of physically distinct channel-precoder pairs, eliminating phase-induced redundancies and thereby leading to significantly improved training efficiency and precision. The proposed framework is independent of the specific precoding algorithm and is applicable to any phase-invariant MU-MISO objective. However, in this work we demonstrate its efficacy using WMMSE-generated labels for WSR maximization.

It is worth highlighting that while the works in~\cite{kim2020deep, zhang2022deep, shi2023robust, shi2021deep} partially mitigate the phase-invariance issue in the output representation by exploiting objective-specific precoder structures, redundancy in the input features remains. In contrast, the proposed framework removes phase-related ambiguities in both the input and output representations and generalizes across a wide range of phase-invariant precoding tasks.

Our contributions are summarized as follows:
\begin{itemize}
    \item \textbf{Geometry-aware learning:} We introduce a DL-based MU-MISO precoding framework in which both channel and precoder vectors for each UE are mapped to points in $\mathbb {CP}^n$, thereby eliminating phase-related ambiguities.
    \item \textbf{Two $\mathbb {CP}^n$ parameterizations are developed:} One uses real-valued embeddings, while the other uses complex hyperspherical coordinates.
    \item \textbf{End-to-end constrained learning:} We decompose each precoder into direction in $\mathbb {CP}^n$ and power allocation factor, and train the DNN using sum-rate-aware loss over a wide SNR range.
    \item \textbf{Benchmarks:} The proposed parameterizations are benchmarked against two baselines, and demonstrate significant performance and generalization gains. 
\end{itemize}
The rest of this paper is summarized as follows. Section~\ref{sec:system} introduces the system model. In Section~\ref{sec:wmmse}, the WSR maximization problem is formulated and the WMMSE algorithm is explained. Section~\ref{sec:cps} provides the required preliminaries on CPSs. The proposed DL framework is introduced and explained in Section~\ref{sec:dnn}. Results and discussions are in Section~\ref{sec:results}. Conclusions are given in Section~\ref{sec:conclusions}. 

\textit{Notations:} Matrices and vectors are denoted by boldface upper-case and lower-case letters, respectively. The $n$th column of $\mathbf X$ is $\mathbf x_n$, and $x_{m,n}$ is the $(m,n)$th element of $\mathbf X$. The $n$th element of $\mathbf x$ is $x_n$. The Euclidean norm of $\mathbf x$ is $\left\| \mathbf x \right\|$. The trace of $\mathbf X$ is $\text{Tr}\left(\mathbf X\right)$. The $n$-dimensional all-zero vector is $\mathbf 0_n$. The conjugate, the Hermitian transpose, and the expectation operators are $(\cdot)^*$, $(\cdot)^H$, and $\mathbb E\{\cdot\}$, respectively. The angle of a complex argument is $\angle{(\cdot)}$. The notations $\mathbb R$, $\mathbb C$, $\mathbb R^n$, $\mathbb C^{n}$, and $\mathbb {CP}^n$ denote respectively the sets of real numbers, complex numbers, $n-$dimensional real vector spaces, $n-$dimensional complex vector spaces, and $n-$dimensional complex projective spaces.
\section{System Model}\label{sec:system}
We consider a single-cell MU-MISO downlink transmission, where a BS with $N$ antennas communicates with $K$ single-antenna UEs using the same time-frequency resources. Frequency-flat Rayleigh fading channels are considered, and the BS is assumed to have perfect CSI knowledge.\footnote{We focus on an ideal scenario here to highlight the gains obtained purely from the proposed CPS-based parameterizations, while the impact of channel and hardware impairments can be investigated in future research.}

Let $\mathbf H = [\mathbf h_1, \cdots, \mathbf h_K]\in \mathbb C^{N\times K}$ be the channel matrix between the BS and all UEs, and $\mathbf W = [\mathbf w_1, \cdots, \mathbf w_K]\in \mathbb C^{N\times K}$ be the corresponding precoder weights constrained by a total power budget $P$ such that $\sum_{k=1}^{K}\left\|\mathbf w_k\right\|^2 \le P$.
The signal received by the $k$th UE is:
\begin{equation}
    y_k = \mathbf{h}_k^H \mathbf{w}_k s_k + \sum_{j=1,\ j\ne k}^{K} \mathbf h_k^H\mathbf w_j s_j + n_k,
\end{equation}where $s_k$ is the information symbol intended for the $k$th UE with $\mathbb E\{|s_k|^2\}=1$, and $n_k$ is the additive white Gaussian noise with zero mean and variance $\sigma^2$. The signal-to-interference-and-noise ratio (SINR) for the same UE is:
\begin{equation}\label{eq:sinr}
    \mathrm{SINR}_k = \frac{\left|\mathbf{h}_k^H \mathbf{w}_k\right|^2}{\sum_{j=1,\ j\ne k}^{K} \left|\mathbf h_k^H\mathbf w_j\right|^2 + \sigma^2}.
\end{equation}
In addition, the achievable sum-rate (bit/s/Hz) is given as:
\begin{equation}
    \mathcal R = \sum_{k=1}^{K} \log_2\left(1 + \mathrm{SINR_k}\right). 
\end{equation}

Next, we formulate the precoding design optimization problem for WSR maximization.
\section{Weighted Sum-Rate Maximization Precoding}\label{sec:wmmse}
 In this work, we consider the WSR maximization as the utility function to be maximized. The corresponding optimization problem is given as follows:
\begin{align} \label{OP1}
	& \hspace{.0cm}\underset{\mathbf W}{\text{maximize}} \hspace{0.25cm}  \sum_{k=1}^{K} \alpha_{k} \log_2\left(1 + \mathrm{SINR_k}\right) \\ 
	&\hspace{0cm}\text{subject to} \nonumber \\ 
	& \hspace{0cm} \text{Tr}\left(\mathbf W \mathbf W^H\right) \le P,
\end{align}where $\alpha_k$ is the priority of the $k$th UE,
which is assumed to be known by the BS. Problem~(\ref{OP1}) is non-convex and known to be NP-hard. To tackle this challenge, the authors in~\cite{shi2011iteratively} demonstrated that problem (\ref{OP1}) is equivalent to the following WMMSE problem in the sense that solving both optimization problems leads to the same optimal solution:
\begin{align} \label{OP2}
	 & \hspace{-.01cm}\underset{\mathbf W,\ \mathbf c,\ \mathbf q}{\text{minimize}} \hspace{0.5cm} f\left(\mathbf W, \mathbf c, \mathbf q\right) = \sum_{k=1}^{K} \alpha_{k} \Big(c_k e_k \hspace{0cm} - \log (c_k)\Big) \\ 
	&\hspace{0cm}\text{subject to} \nonumber \\ 
	& \hspace{0cm} \text{Tr}\left(\mathbf W \mathbf W^H\right) \le P, \label{eq: power constraint}
\end{align}where $\mathbf c = \left[c_{1}, \cdots, c_K\right]$ with $c_k$ being a positive weight parameter assigned to the $k$th UE, $\mathbf q = [q_{1}, \cdots, q_K]$ with $q_k$ being the filter weight applied at the $k$th UE to recover its intended symbol $s_k$, and $e_k$ reflects the mean square error (MSE) at the $k$th UE. 

Problem (\ref{OP2}) is not jointly convex in $\{\mathbf W, \mathbf q, \mathbf c\}$. However, it is convex in each optimization variable separately. The WMMSE algorithm~\cite{shi2011iteratively} solves (\ref{OP2}) iteratively and is guaranteed to converge to a stationary point of the WSR maximization problem (\ref{OP1}). 
\subsection{WMMSE Algorithm}
In this section, we provide a brief summary of the main steps of the WMMSE algorithm, which will be used to generate labels for model training. For a detailed implementation of the WMMSE algorithm, we refer the reader to \cite{shi2011iteratively, abdullah2025impact}.

A single WMMSE iteration involves updating the following three different parameters in an alternating fashion: 
\paragraph{Receive Filter $(q_k^{\star})$}For the $k$th UE, the minimum mean squared error (MMSE) filter is:
\begin{equation}\label{eq: q_star}
    q_k^{\star} = \frac{\mathbf w_k^H\mathbf h_k}{\sum_{j=1}^{K}\left|\mathbf h_k^H\mathbf w_{j}\right|^2 + \sigma^2},
\end{equation}where the precoders are initialized to any feasible point.
\paragraph{Weight Parameter $(c_k^\star)$}The weight assigned to the $k$th UE depends on the MSE of the received signal, given as:
\begin{align} \label{eq: e_star}
   e_k & = \left|q_k \mathbf h_k^H \mathbf w_k - 1\right|^2  + \sum_{j=1,\ j\neq k}^K \Big|q_k \mathbf h_k^H\mathbf w_j\Big|^2 + \left|q_k\right|^2 \sigma^2. 
\end{align}The optimal weight is then obtained as $c_k^{\star} = e_k^{-1}$. 
\paragraph{Precoder Weights $(\mathbf w_k^\star)$}The final step is finding the precoding weights for each UE. This step requires forming a Lagrangian function and applying first-order optimality conditions. The solution for the $k$th UE precoder is given as:
\begin{equation} \label{eq: w_star}
    \mathbf w_k^\star = \alpha_k c_k q_k^\ast \Big(\mathbf G + \lambda \mathbf I_N\Big)^{-1} \mathbf h_k,
\end{equation}where $\mathbf I_N$ is the $N\times N$ identity matrix, $\lambda \ge 0$ is the Lagrangian multiplier, and
\begin{equation}
    \mathbf G = \sum_{k=1}^{K} \alpha_k c_k \big|q_k\big|^2 \mathbf h_k \mathbf h_k^H. 
\end{equation}To satisfy the complementary slackness condition, either $\lambda=0$ or $\sum_{k=1}^{K}\left\|\mathbf w_k\right\|^2=P$ must hold. If $\lambda = 0$ violates the power constraint, an optimal $\lambda$ must be calculated (e.g., via bisection method) that satisfies the power constraint with strict equality. 

The WMMSE solution repeats steps (a)-(c) iteratively until convergence. We summaraize the main steps in Algorithm~\ref{alg: wmmse}. 

Next, we present the preliminaries on CPSs.
\begin{algorithm}[t]
	\caption{WMMSE Algorithm \cite{shi2011iteratively}}
	\label{alg: wmmse}
	\begin{algorithmic}[1]
		\STATE \textbf{input:} $\mathbf H$, feasible $\mathbf W$, $P$, $\sigma^2$, $\alpha_k\ (\forall k)$, $\mathrm{MAX\_ITER}$
		\STATE \textbf{for} $iter = 1 \rightarrow \mathrm{MAX\_ITER}$
		\STATE \ \ \ \ \ \textbf{calculate} $q_k^{\star}$ according to (\ref{eq: q_star})
        \STATE \ \ \ \ \ \textbf{calculate} $e_k$ from (\ref{eq: e_star}) and substitute $c_k^\star = e_k^{-1}$ 
        \STATE \ \ \ \ \ \textbf{calculate} $\mathbf w_k^\star$ according to (\ref{eq: w_star}) for $\lambda = 0$
        \STATE \ \ \ \ \ \textbf{if} $\sum_{k=1}^{K}\left\|\mathbf w_k^\star\right\|^2 > P$
        \STATE \ \ \ \ \ \ \ \ \ \textbf{apply} bisection method to find $\lambda^\star$ 
        \STATE \ \ \ \ \ \ \ \ \ \textbf{re-calculate} $\mathbf w_k^\star$ \vspace{0cm}
        \STATE \ \ \ \ \ \textbf{if converged}: \textbf{break}
        \STATE \textbf{end for}
		\STATE \textbf{output:} $\mathbf w_k^\star\ (\forall k)$
	\end{algorithmic}
\end{algorithm}
\section{Preliminaries on Complex Projective Spaces}\label{sec:cps}
\subsection{Definition}
An $n$-dimensional complex projective space $\mathbb {CP}^n$ is the space of complex lines in $\mathbb C^{n+1}$ passing through the origin.\footnote{It is worth noting that $\mathbb {CP}^1$ is the Riemann sphere.} In $\mathbb {CP}^{n}$, any two vectors $\{\mathbf v, \mathbf u\} \in \mathbb C^{n+1}$ differing by a non-zero complex scaling are equivalent. Specifically, if $\mathbf v = \lambda \mathbf u$ for some non-zero $\lambda \in \mathbb C$ and $\mathbf v \ne \mathbf 0_{n+1}$, then $\mathbf v$ and $\mathbf u$ are the same point in $\mathbb {CP}^n$~\cite{bengtsson2017geometry}. Moreover, if $\{\bar{\mathbf v}, \bar{\mathbf u}\} \in\mathbb C^{n+1}$ are unit-norm vectors such that $\left\| \bar{\mathbf v}\right\| = \left\| \bar{\mathbf u}\right\| = 1$, the equivalence relation in $\mathbb{CP}^n$ reduces to a global phase rotation. That is, $\bar{\mathbf v}$ and $\bar{\mathbf u}$ are the same point in $\mathbb {CP}^n$ if $\bar{\mathbf v} = e^{\jmath \psi} \bar{\mathbf u}$ for any $\psi \in\mathbb R$.  
\subsection{Embeddings and Dimensionality}
A unit-norm complex-valued vector ${\mathbf u} \in\mathbb C^{n+1}$ possesses $2n+1$ degrees of freedom (DoF), as the unit-norm constraint reduces the original (unconstrained) $2n+2$ DoF by one. If the global phase is further removed, the resulting vector $\widetilde{\mathbf u}$ is constrained to a representation with exactly $2n$ DoF, which is the intrinsic dimensionality of $\mathbb {CP}^n$. 

While a standard real-valued embedding of an element in $\mathbb {CP}^n$ still necessitates $2n+1$ coordinates ($n+1$ real components and $n$ non-zero imaginary components), $\widetilde{\mathbf{u}}$ can be parameterized with exactly $2n$ parameters using complex hyperspherical coordinates as detailed next.
\subsection{Complex Hyperspherical Coordinates}
A unit-length vector $\mathbf{u} \in \mathbb{C}^{n+1}$ can be parameterized via complex hyperspherical coordinates $\{\boldsymbol{\theta}, \boldsymbol{\phi}\}\in \mathbb R^{n}$ as follows:
\begin{equation}\label{eq:hsc}
\begin{bmatrix}
 u_1\\
 u_2\\
 \vdots\\
 u_{n}\\
u_{n+1}
\end{bmatrix} = e^{\jmath \phi_0}
\begin{bmatrix}
\cos{\theta_1}  \\
 e^{\jmath \phi_1} \cos{\theta_2} \sin{\theta_1} \\
 \vdots \\
 e^{\jmath \phi_{n-1}}  \cos{\theta_{n}}\prod_{i=1}^{n-1} \sin{\theta_i} \\
e^{\jmath \phi_{n}} \prod_{i=1}^{n} \sin{\theta_i}
\end{bmatrix},
\end{equation}
where $(\phi_1, \cdots, \phi_n) \in [-\pi, \pi]$ are the phase angles and $(\theta_1, \cdots, \theta_n) \in [0, \pi/2]$ are the amplitude angles. The term $\phi_0$ represents the global phase. In the context of $\mathbb{CP}^n$, $\phi_0$ is arbitrary and can be omitted, leaving the vector fully determined by the $2n$ parameters $(\theta_1, \dots, \theta_n, \phi_1, \dots, \phi_n)$.

The phase angles of $\mathbf u$, and assuming that its global phase is removed, can be obtained as follows: 
\begin{equation}
    \phi_m = \angle(u_{m+1}), \hspace{0.5cm} \forall m \in \{1, \cdots, n\}.
\end{equation}
The amplitude angles can be obtained as:
\begin{equation}\label{eq:theta}
    \theta_m = \arctan{\left(\frac{\left\|\mathbf u_{m+1:}\right\|}{|u_m|}\right)}, \hspace{0.5cm} \forall m\in\{1, \cdots, n\}
\end{equation}with $\left\|\mathbf u_{l:}\right\| = \sqrt{\sum_{i=l}^{n+1}|u_i|^2}$. \textit{See the Appendix for the proof.} 

Next, we present in detail the proposed DL-based MU-MISO precoding in $\mathbb {CP}^n$.
\section{Neural Precoding Design in  \texorpdfstring{$\mathbb{CP}^n$}{CP	extsuperscript{n}}}\label{sec:dnn}
\subsection{Motivation}
As can be seen from the SINR expression in~(\ref{eq:sinr}), the precoding performance depends on the magnitudes of inner products between channel and precoding vectors, which are invariant to global phase rotations since $\left|\mathbf h_k^H \mathbf w_l\right| = \left|\mathbf h_k^H (e^{\jmath \psi} \mathbf w_l) \right| = \left|(e^{\jmath \psi}\mathbf h_k)^H \mathbf w_l \right|$ for any $\psi \in \mathbb R$. Consequently, neither the channel nor the precoder is uniquely defined, and only their equivalence classes under global phase rotations are physically meaningful. These equivalence classes are naturally represented as points in $\mathbb{CP}^n$.

As a result, conventional parameterizations that rely on real/imaginary or amplitude/phase representations force the DNN to implicitly learn entire equivalence classes of phase-rotated vectors, introducing unnecessary redundancy and increasing the difficulty of the learning task. In contrast, by parameterizing both $\mathbf h_k$ and $\mathbf w_k$ directly in $\mathbb{CP}^n$, the proposed framework operates on geometrically consistent representations, eliminating phase-induced ambiguity and improving learning efficiency. 

Moreover, normalizing $\mathbf h_k$ and $\mathbf w_k$ to be unit-norm enables hyperspherical coordinate parameterization, and it also naturally decomposes the learning problem into two components: (i) learning the precoder shape (direction in $\mathbb {CP}^n$), and (ii) learning the power allocation across UEs.
\subsection{$\mathbb {CP}^n$ Projection \& Parameterizations}
We propose two CPS-based parameterizations. In both parameterizations, channel and precoding vectors for each UE $\{\mathbf h_k, \mathbf w_k\}\in\mathbb C^{N}$ are mapped to $\mathbb {CP}^{N-1}$. 

Specifically, we normalize each channel and precoding vector, $\mathbf h_k$ and $\mathbf w_k$, to have a unit length and remove its global phase using the first element as a phase reference. The resulting projected channel and precoding vectors, ${\mathbf g}_k$ and ${\mathbf u}_k$, represent points in $\mathbb {CP}^{N-1}$ with $2N-2$ DoF. For example, $\mathbf g_k$ is obtained as follows:
\begin{equation}
    \mathbf g_k = \frac{\mathbf h_k}{\left\|\mathbf h_k\right\|} e^{-\jmath\angle{h_{k,1}}}.
\end{equation}In addition, the magnitudes of $\left\|\mathbf h_k\right\|$ and $\left\|\mathbf w_k\right\|$ are preserved to maintain channel strength and power allocation information.

The first parameterization proposed, denoted as $\mathcal P_{cps}$, characterizes these projected vectors ($\mathbf g_k$ and $\mathbf u_k$) using real-valued embeddings. The second parameterization proposed, denoted as $\mathcal P_{hsc}$, utilizes complex hyperspherical coordinates to characterize ${\mathbf g}_k$ and ${\mathbf u}_k$ as will be explained in Section~\ref{sec:preprocessing}.

The proposed parameterizations are compared with two benchmarks. The first benchmark is the classical real/imaginary parameterization, denoted as $\mathcal P_{ri}$, where each channel and precoding vector is described using its real and imaginary components after scaling. The second benchmark, denoted as $\mathcal P_{ncv}$, uses embeddings of normalized complex vectors without removing the global phase. Specifically, each channel and precoder vector is normalized to unit length and parameterized using its real and imaginary components.
\subsection{Raw Dataset Generation}
We generate $10^6$ samples of $N \times K$ MU-MISO channels. For each sample, the WMMSE precoder is computed via Algorithm~\ref{alg: wmmse} with an SNR ($P/\sigma^2$) sampled uniformly between $[0, 20]$ dB. The data samples are shuffled and split into training $(70\%)$, validation $(15\%)$, and testing $(15\%)$.
\subsection{Preprocessing}\label{sec:preprocessing}
In the following, we provide the preprocessing steps applied in each parameterization. 
\begin{table*}[]
\caption{Summary of Number of Input \& Output Features.}
\label{tab:parameterizations}
\centering
\begin{tabular}{|c|c|c|c|}
\hline
\multirow{2}{*}{\textbf{Parameterization}} & \multirow{2}{*}{\textbf{Number of input features}} & \multirow{2}{*}{\textbf{Number of labels}} & \multirow{2}{*}{\textbf{Global phase removed}} \\
                                           &                                          &                                           &                                                \\ \hline
\multirow{2}{*}{$\mathcal P_{ri}$ (benchmark)}         & \multirow{2}{*}{$2NK+1$}                 & \multirow{2}{*}{$2NK$}                    & \multirow{2}{*}{No}                            \\
                                           &                                          &                                           &                                                \\ \hline
\multirow{2}{*}{$\mathcal P_{ncv}$ (benchmark)}        & \multirow{2}{*}{$2NK+K+1$}               & \multirow{2}{*}{$2NK+K$}                  & \multirow{2}{*}{No}                            \\
                                           &                                          &                                           &                                                \\ \hline
\multirow{2}{*}{$\mathcal P_{cps}$ (proposed)}        & \multirow{2}{*}{$2NK+1$}                 & \multirow{2}{*}{$2NK$}                    & \multirow{2}{*}{Yes}                           \\
                                           &                                          &                                           &                                                \\ \hline
\multirow{2}{*}{$\mathcal P_{hsc}$ (proposed)}        & \multirow{2}{*}{$K(3N-2)+1$}             & \multirow{2}{*}{$K(3N-2)$}                & \multirow{2}{*}{Yes}                           \\
                                           &                                          &                                           &                                                \\ \hline
\end{tabular}
\end{table*}
\subsubsection{Parameterization \texorpdfstring{$\mathcal P_{cps}$}{(Pcps)}} The projected channel vector to $\mathbb {CP}^{N-1}$, denoted as ${\mathbf g}_k$ for the $k$th UE, is parameterized with $2N-1$ real-valued coefficients ($N$ real and $N-1$ imaginary components). The channel norms $\left\|\mathbf h_k\right\|$ and the SNR are scaled to $[-1, 1]$. For the channel norms, the scaling is performed using exclusively training data information. The number of input features is $2NK+1$, comprising $K(2N-1)$ for ${\mathbf g}_k(\forall k)$, $K$ scaled channel norms, and the scaled SNR. 

Similarly, the projected precoding vector to $\mathbb {CP}^{N-1}$, denoted as ${\mathbf u}_k$ for the $k$th UE, is parameterized by $2N-1$ real coefficients. The power allocation ratio $\zeta_k$ for each UE is then derived from the vector norms $\left\|\mathbf w_k\right\|$. These power ratios satisfy $\sum_{k=1}^{K}\zeta_k=1$. This results in $2NK$ output labels: $K(2N-1)$ for the precoding direction $\mathbf u_k (\forall k)$ and $K$ for the power ratios.
\subsubsection{Parameterization \texorpdfstring{$\mathcal P_{hsc}$}{(Phsc)}} Here, complex hyperspherical coordinates are utilized to parameterize $\mathbf g_k$. We denote the amplitude angles of $\mathbf g_k$ as $\boldsymbol \theta_{g_k}$ and the phase angles as $\boldsymbol \phi_{g_k}$. A term of $\pi/4$ is subtracted from each of the $N-1$ amplitude angles $(\boldsymbol \theta_{g_k})$ so that these features are centered around $0$. Furthermore, each of the $N-1$ phase angles $(\boldsymbol \phi_{g_k})$ is encoded using two terms, $\sin{\phi}$ and $\cos{\phi}$, to remove the discontinuity around $\pm \pi$. The total number of input features is $K(3N-2)+1$ divided as: $K(N-1)$ for ${\boldsymbol \Theta_g} = [{\boldsymbol \theta}_{g_1}, \cdots, {\boldsymbol \theta}_{g_K}]\in\mathbb R^{N-1\times K}$, $K(2N-2)$ for encoding ${\boldsymbol \Phi_g} = [{\boldsymbol \phi}_{g_1}, \cdots, {\boldsymbol \phi}_{g_K}]\in\mathbb R^{N-1\times K}$, $K$ for scaled channel norms, and $1$ for scaled SNR. 

Similarly, ${\mathbf u}_k$ is parameterized by $\{\boldsymbol \theta_{u_k},\boldsymbol \phi_{u_k}\}$, and the total number of labels is $K(3N-2)$, divided as: $K(N-1)$ for ${\boldsymbol \Theta_u} = [{\boldsymbol \theta}_{u_1}, \cdots, {\boldsymbol \theta}_{u_K}]\in\mathbb R^{N-1\times K}$, $K(2N-2)$ for encoding ${\boldsymbol \Phi_u} = [{\boldsymbol \phi}_{u_1}, \cdots, {\boldsymbol \phi}_{u_K}]\in\mathbb R^{N-1\times K}$, and $K$ for the power split ratios among UEs.
\subsubsection{Parameterization \texorpdfstring{$\mathcal P_{ncv}$}{(Pncv)} (benchmark)}The normalized channel vectors $\mathbf a_k = \frac{\mathbf h_k}{\left\|\mathbf h_k\right\|} \in \mathbb C^{N}$, and the normalized precoding vectors $\mathbf b_k = \frac{\mathbf w_k}{\left\|\mathbf w_k\right\|} \in \mathbb C^{N}$ are parameterized using their real and imaginary components. The channel norms and the SNR are scaled to $[-1, 1]$. The number of input features is $2NK+K+1$ divided as: $2NK$ for $\mathbf a_k (\forall k)$, $K$ scaled channel norms, and $1$ for scaled SNR. 

The number of labels is $2NK+K$ comprising $2NK$ for the real and imaginary components of $\mathbf b_k (\forall k)$, and $K$ for the power ratios of all UEs. 
\subsubsection{Parameterization \texorpdfstring{$\mathcal P_{ri}$}{(Pri)} (benchmark)} Channel vectors $\mathbf h_k$ and precoding vectors $\mathbf w_k$ are first scaled to $[-1, 1]$ using training data information, and then parameterized using their real and imaginary components. The number of input features is $2NK+1$, comprising $2NK$ for the real and imaginary parts of the scaled $\mathbf h_k (\forall k)$ and $1$ for the scaled SNR. The number of labels is $2NK$ representing the real and imaginary components of the scaled $\mathbf w_k (\forall k)$. 

Table~\ref{tab:parameterizations} summarizes the number of input features and labels for each parameterization.
\subsection{DNN Architecture} The DNN considered is a classical feed-forward DNN.\footnote{Our aim here is to highlight the gains obtained purely from the CPS-based parameterizations, rather than from architectural choices. Nonetheless, designing DL models that exploit the geometry of $\mathbb {CP}^{n}$ in the network architecture is an interesting direction for future research.} The input layer has a dimension equal to the number of input features for each parameterization. This layer is followed by $5$ fully connected layers. The first $4$ are hidden layers with $\{128, 256, 256, 128\}$ neurons, respectively. The first three hidden layers deploy Batch Normalization (BN) before the activations. The parametric rectified linear unit (PReLU) activations are applied in all hidden layers. The output layer uses linear activation with no BN, and its number of neurons matches the number of labels for each parameterization. 

The number of trainable parameters is $141,476$ for both $\mathcal P_{ri}$ and $\mathcal P_{cps}$, while for $\mathcal P_{ncv}$ and $\mathcal P_{hsc}$, the parameter counts are $142,504$ and $143,532$, respectively.
\subsection{Post-Processing}
Differentiable postprocessing layers are implemented after the output layer for each parameterization except $\mathcal P_{ri}$, as detailed in the following paragraphs. 

For $\mathcal P_{cps}$, the model constructs $\widetilde{\mathbf U} = [\widetilde{\mathbf u}_1, \cdots, \widetilde{\mathbf u}_K]\in\mathbb C^{N\times K}$ from the first $K(2N-1)$ DNN outputs. These vectors serve as canonical representatives of points in $\mathbb {CP}^{N-1}$, as each vector is constrained to unit length with a zero-phase first element. Further, a $\mathrm{softmax}$ layer is applied for the final $K$ DNN outputs to obtain the predicted power split ratios $\tilde \zeta_k (\forall k)$. 

For $\mathcal P_{hsc}$, a $\tanh$ layer is applied for the first $K(N-1)$ outputs to restore the amplitude angles $\widetilde{\boldsymbol \Theta}_u = [\widetilde{\boldsymbol \theta}_{u_1}, \cdots, \widetilde{\boldsymbol \theta}_{u_K}]\in\mathbb R^{N-1\times K}$. The angles are adjusted to match the original range of $[0, \pi/2]$. The following $K(2N-2)$ outputs are reshaped, normalized, and processed to obtain the predicted phase angles $\widetilde{\boldsymbol \Phi}_u = [\widetilde{\boldsymbol \phi}_{u_1}, \cdots, \widetilde{\boldsymbol \phi}_{u_K}]\in\mathbb R^{N-1\times K}$ in $[-\pi, \pi]$. A $\mathrm{softmax}$ layer is applied on the last $K$ DNN outputs to obtain $\tilde{\zeta}_k(\forall k)$. 

For $\mathcal P_{ncv}$, the model constructs $\widetilde{\mathbf B} = [\widetilde{\mathbf b}_1, \cdots, \widetilde{\mathbf b}_K]\in\mathbb C^{N\times K}$ from the first $2KN$ DNN outputs. Each of the $K$ vectors in $\widetilde{\mathbf B}$ has unit length. Unlike $\mathcal P_{cps}$, these vectors do not uniquely represent points in $\mathbb {CP}^{N-1}$ because their global phase is not removed. A softmax layer is applied for the final $K$ DNN outputs to obtain the predicted power split ratios. 

In summary, the DNN predictions in both $\mathcal P_{cps}$ and $\mathcal P_{hsc}$ represent phase-invariant precoding directions in $\mathbb {CP}^{N-1}$ along with power ratios. The former uses a direct vector representation, while the latter uses complex hyperspherical coordinates. In contrast, the benchmark parameterization $\mathcal P_{ncv}$ predicts power ratios and normalized precoding vectors on the unit sphere, which inherently include a global phase. 
\begin{figure*}[t]
    \centering
    \includegraphics[width=17.0cm,height=6.0cm]{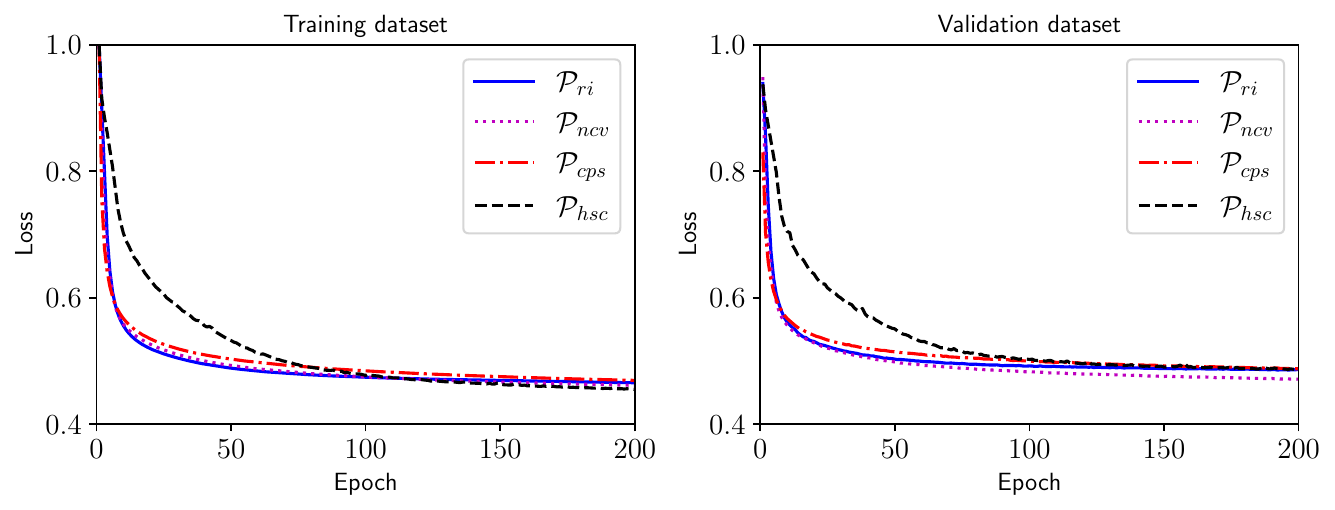}
    \caption{Model convergence of different parameterizations.}
    \label{fig:converg_loss}
\end{figure*}
\subsection{Loss Functions}
The following losses are implemented during training:
\paragraph{Power Split Loss} As the power ratios sum to one and are non-negative, they are treated as probabilities. The log-based KL (Kullback–Leibler) divergence is utilized to quantify the loss between the true ($\zeta_k$) and predicted ($\tilde \zeta_k$) power split: 
\begin{equation}
    \mathcal L_{\zeta}^{(i)} = \sum_{k=1}^{K} \zeta_k^{(i)} \log\left(\zeta_k^{(i)}\right) - \sum_{k=1}^{K}\zeta_k^{(i)}\log\left(\tilde{\zeta}_k^{(i)}\right),
\end{equation}where $i$ is the sample index.
\paragraph{Precoder Direction Loss} The loss between the normalized WMMSE precoder $\mathbf z_k\in\{\mathbf u_k, \mathbf b_k\}$ and the unit-norm DNN predicted precoder $\widetilde{\mathbf z}_k\in\{\widetilde{\mathbf u}_k, \widetilde{\mathbf b}_k\}$ is measured as follows:\footnote{Note that the precoder direction loss is applied in both $\mathcal P_{cps}$ and $\mathcal P_{ncv}$ parameterizations, since the precoding vectors predicted by the DNN have unit length in both cases.}
\begin{equation}\label{eq:dir}
    \mathcal L_{z}^{(i)} = \frac{1}{K}\sum_{k=1}^{K}\left(1 - \left|(\mathbf z_k^{(i)})^H\widetilde{\mathbf z}_k^{(i)}\right|^2\right).
\end{equation}
\paragraph{Angles Loss}The loss between the true angles $\boldsymbol \Psi \in \{\boldsymbol \Theta_u, \boldsymbol \Phi_u\}$ and the predicted ones $\widetilde {\boldsymbol \Psi} \in \{\widetilde{\boldsymbol \Theta}_u, \widetilde{\boldsymbol \Phi}_u\}$ is quantified by measuring the circular phase distance as follows:
\begin{equation}
    \mathcal L_{\Psi}^{(i)} = \frac{1}{K(N-1)}\sum_{k=1}^{K}\sum_{n=1}^{N-1}\left(1 - \cos{\left(\psi_{k,n}^{(i)} - \widetilde \psi_{k,n}^{(i)}\right)}\right).
\end{equation}
\subsection{Total Losses} The total losses for each parameterization are as follows.
\paragraph{Parameterization \texorpdfstring{$\mathcal P_{cps}$}{Pcps}}The total loss for $\mathcal P_{cps}$ is the sum of the power split loss and the precoder direction loss, given as follows:
\begin{equation}
   \mathcal L_{cps} = \frac{\sum_{i=1}^{I} \beta^{(i)}(\mathcal L_\zeta^{(i)} + \mathcal L_u^{(i)})}{\sum_{i=1}^{I}\beta^{(i)}}
\end{equation}with $I$ being the number of samples, $\beta^{(i)}$ is a rate penalty defined as $\beta^{(i)}={\mathcal R}^{(i)} / \widetilde{\mathcal R}^{(i)}$ with $\mathcal R^{(i)}$ and $\widetilde{\mathcal R}^{(i)}$ being the sum-rates achieved by the WMMSE algorithm and the DNN-predicted precoders, respectively, and $\mathcal L_u$ denotes the precoder direction loss in (\ref{eq:dir}) evaluated for $\mathbf{z} = \mathbf{u}$.
\paragraph{Parameterization \texorpdfstring{$\mathcal P_{hsc}$}{Phsc}}The total loss of $\mathcal P_{hsc}$ is the sum of the power split loss and the angles losses given as:
\begin{equation}
   \mathcal L_{hsc} = \frac{\sum_{i=1}^{I} \beta^{(i)}(\mathcal L_\zeta^{(i)} + \mathcal L_{\Theta_u}^{(i)} + 0.5\mathcal L_{\Phi_u}^{(i)})}{\sum_{i=1}^{I}\beta^{(i)}}.
\end{equation}The scaling by $0.5$ ensures that amplitude- and phase-angles losses have a similar range.
\paragraph{Parameterization $\mathcal P_{ncv}$} The total loss for this benchmark method is the sum of the power split loss and the precoder direction loss, given as:
\begin{equation}
    \mathcal L_{ncv} = \frac{\sum_{i=1}^{I}\beta^{(i)}(\mathcal L_{\zeta}^{(i)} + \mathcal L_b^{(i)})}{\sum_{i=1}^{I}\beta^{(i)}},
\end{equation}where $\mathcal L_b$ is the loss in (\ref{eq:dir}) evaluated for $\mathbf{z} = \mathbf{b}$.
\paragraph{Parameterization \texorpdfstring{$\mathcal P_{ri}$}{Pri}} For the classical real/imaginary benchmark parameterization, we apply the standard MSE loss weighted by the rate penalty
\begin{equation}
    \mathcal L_{ri} = \frac{\sum_{i=1}^{I} \beta^{(i)} \left\|\mathbf w_{true}^{(i)} - \widetilde{\mathbf w}_{pred}^{(i)}\right\|^2}{2NK \sum_{i=1}^{I}\beta^{(i)}},
\end{equation}where $\mathbf w_{true}\in\mathbb R^{2NK}$ contains the true WMMSE coefficients after scaling, and $\widetilde{\mathbf w}_{pred}\in\mathbb R^{2NK}$ is a vector containing the raw DNN predictions.

Next, we present and discuss the results.
\section{Results and Discussions} \label{sec:results}
\subsection{Simulation \& Training Parameters}
The system parameters are as follows: The number of BS antennas $N=4$, the number of UEs $K=4$, and the noise variance $\sigma^2=1$. For the WMMSE algorithm, the number of iterations is $10$ and the UE priority $\alpha_k=1 (\forall k)$. 

The DNN parameters are selected, after fine-tuning, as follows: The learning rate is $10^{-3}$, the batch size is $1024$, the number of epochs is $200$, and the Adam (adaptive
moment estimation) algorithm~\cite{kingma2014adam} is used to update the network parameters via backpropagation. 

Data generation and preprocessing are performed on Intel(R) Core(TM) Ultra 7 165H central processing unit (CPU) with 64 GB of RAM (random access memory). The DNN models are trained using PyTorch on NVIDIA RTX 1000 Ada Generation Laptop GPU (graphics processing unit) that has 6 GB of dedicated memory. 

The documented code used for the full training pipeline is available at~\cite{github_mimo_cps} for reproducibility.
\begin{figure*}[t]
    \centering
    \includegraphics[width=17.0cm,height=6.0cm]{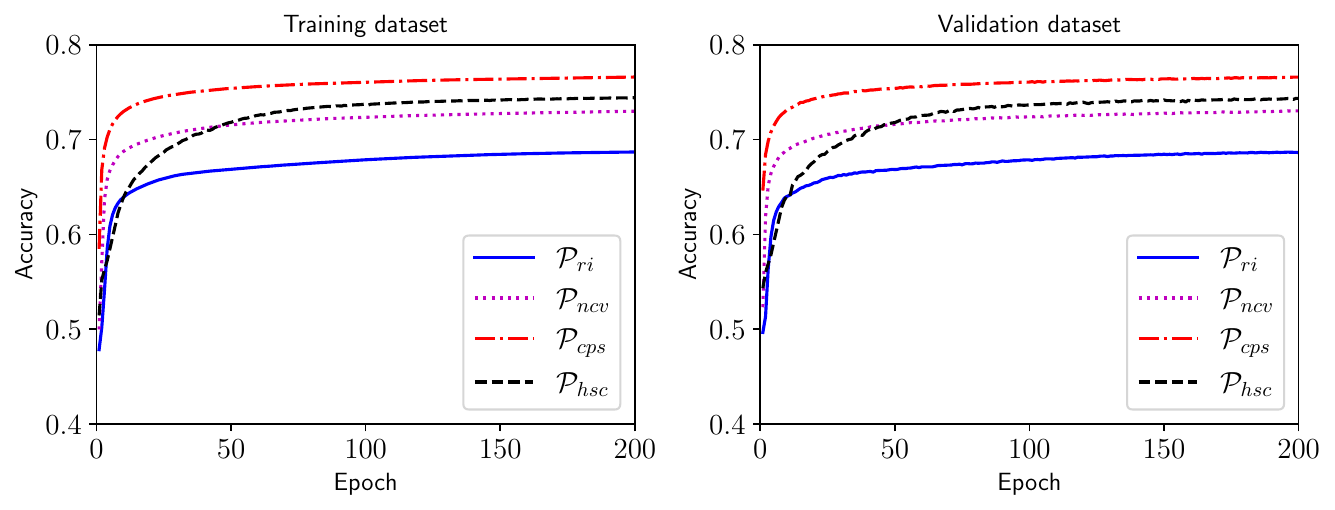}
    \caption{Accuracy vs epoch for different parameterizations.}
    \label{fig:converg_acc}
\end{figure*}
\subsection{Convergence Analysis}
In Fig.~\ref{fig:converg_loss}, we illustrate the convergence behavior in terms of the total loss of different parameterizations.\footnote{Since different parameterizations have different loss ranges, the curves in Fig.~\ref{fig:converg_loss} are normalized by the highest training loss. For each parameterization, both training and validation data are normalized by the same factor to ensure that the generalization gap remains proportional and meaningful.} It can be observed that parameterizations $\mathcal P_{ncv}$, $\mathcal P_{cps}$ and $\mathcal P_{ri}$ achieve faster convergence than $\mathcal P_{hsc}$. Importantly, all trained models show no sign of overfitting, as validation losses continue to decrease until convergence. 

Fig.~\ref{fig:converg_acc} shows the convergence behavior in terms of the sum-rate accuracy of different parameterizations.\footnote{We define the accuracy as the sum-rate achieved with DNN-predicted solutions divided by the WMMSE-based sum-rate, averaged over all samples.} The CPS-based parameterizations, and particularly $\mathcal P_{cps}$, demonstrate large gains over the benchmark parameterizations $\mathcal P_{ncv}$ and $\mathcal P_{ri}$. Specifically, at epoch $200$, the benchmark methods have rate accuracies of $68.6\%$ for $\mathcal P_{ri}$ and $73\%$ for $\mathcal P_{ncv}$, while CPS parameterizations $\mathcal P_{hsc}$ and $\mathcal P_{cps}$ achieve $74.4\%$ and $76.6\%$ rate accuracies, respectively.\footnote{The performance was tested with different number of DNN layers, UEs, and antennas. $\mathcal P_{cps}$ consistently outperformed $P_{ri}$ and $P_{ncv}$ by a large margin, while the gain of $\mathcal P_{hsc}$ varied depending on model complexity and system dimensions.} Interestingly, this large performance gain by $\mathcal P_{cps}$ is achieved while requiring the exact same number of input features, labels, and trainable parameters as $\mathcal P_{ri}$. In contrast, the benchmark $\mathcal P_{ncv}$ has a higher parameter count than $\mathcal P_{cps}$ and leads to a lower prediction accuracy. 

Comparing the two CPS-based parameterizations, and despite using a smooth $\sin-\cos$ encoding for relative phases, $\mathcal P_{hsc}$ with hyperspherical coordinates consistently underperforms $\mathcal P_{cps}$ with real-valued embeddings, while incurring higher computational cost (parameter count). The nested trigonometric couplings in hyperspherical coordinates likely create ill-conditioned gradients, making optimization harder in standard feedforward network architectures, despite preserving the exact $\mathbb {CP}^n$ DoF. 

The results in both Fig.~\ref{fig:converg_loss} and Fig.~\ref{fig:converg_acc} are averaged over $10$ independent training sessions with different seeds.
\subsection{Model Generalization}
In this section, we present the model generalization across a wide range of SNRs in terms of the accuracy and the achievable sum-rate using samples from the test dataset. For this purpose, the model that achieved the highest validation accuracy in the $10$ independent training sessions was chosen as the final model for each parameterization.

\begin{figure}[t]
    \centering
    \includegraphics[width=8.50cm,height=6.0cm]{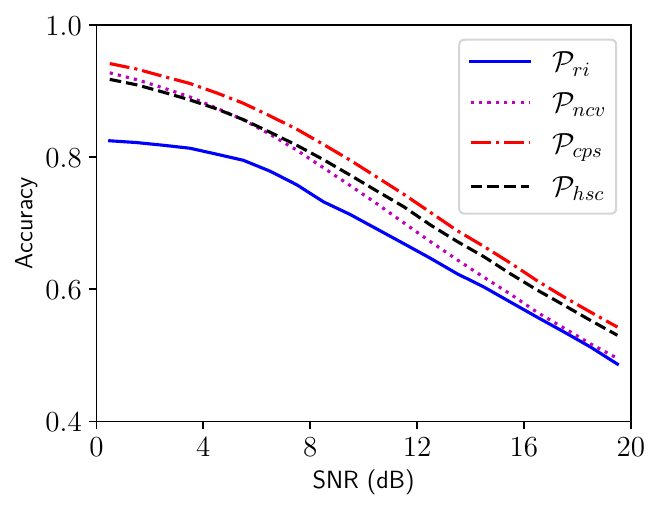}
    \caption{Model generalization: Accuracy vs SNR.}
    \label{fig:acc_generalization}
\end{figure}

In Fig.~\ref{fig:acc_generalization}, we see that parameterization $\mathcal P_{cps}$ achieves the highest accuracy across the entire SNR range, while $\mathcal P_{ri}$ achieves the lowest accuracy, despite the fact that both parameterizations have the exact same number of trainable parameters. The gap between the two parameterizations is particularly large at low SNR levels. For example, at $0$ dB SNR, $\mathcal P_{cps}$ achieves $94\%$ of the WMMSE rate, compared to $82.5\%$ for $\mathcal P_{ri}$. In addition, $\mathcal P_{hsc}$ shows close performance to that of $\mathcal P_{cps}$ for the entire SNR range. Interestingly, $\mathcal P_{ncv}$ shows a comparable performance to the CPS-based parameterizations only at low SNR levels. As the SNR increases, the gap between $\mathcal P_{ncv}$ and both $\mathcal P_{cps}$ and $\mathcal P_{hsc}$ becomes larger. At very high SNRs, the accuracy of the $\mathcal P_{ncv}$ parameterization becomes almost identical to that of $\mathcal P_{ri}$. 

\begin{figure}[t]
    \centering
    \includegraphics[width=8.50cm,height=6.0cm]{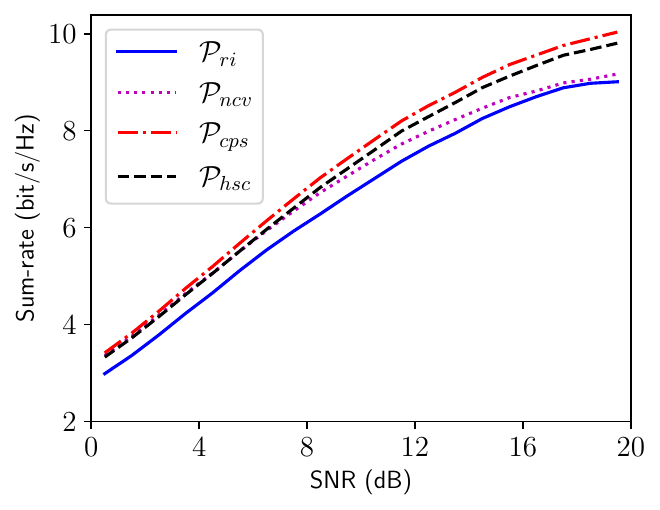}
    \caption{Model generalization: Sum-rate vs SNR.}
    \label{fig:rate_generalization}
\end{figure}

Fig.~\ref{fig:rate_generalization} demonstrates the generalization behavior in terms of the achievable sum-rate. Similar to Fig.~\ref{fig:acc_generalization}, parameterization $\mathcal P_{ri}$ has the worst sum-rate generalization for the entire SNR range. In addition, while $\mathcal P_{ncv}$ shows improved performance compared to $\mathcal P_{ri}$, the CPS-based parameterizations $\mathcal P_{cps}$ and $\mathcal P_{hsc}$ significantly outperform the benchmarks at medium and high SNRs. For example, at SNR of $10$ dB, $\mathcal P_{cps}$ outperforms $\mathcal P_{ncv}$ by $0.4$ bit/s/Hz and $\mathcal P_{ri}$ by $0.8$ bit/s/Hz. This gain further increases to $0.9$ and $1$ bit/s/Hz, respectively, at SNR of $18$ dB. 

Finally, from Fig.~\ref{fig:acc_generalization} and Fig.~\ref{fig:rate_generalization}, we observe that in the low SNR regime, decomposing the learning task into predicting normalized precoding vectors and power split ratios (as in $\mathcal P_{ncv}$) can achieve higher rates compared to the classical approach $\mathcal P_{ri}$, although this improvement comes at the cost of an increased number of model parameters. However, in the medium- and high-SNR regimes, this decomposition alone is no longer sufficient, and the proposed geometry-aware, phase-invariant parameterizations are far more superior, even without necessarily increasing the total DNN parameter count.
\subsection{End-to-End (E2E) Inference Times}
The E2E inference times for the WMMSE algorithm and the DNN-based solutions are shown in Table~\ref{tab:inference}. The DNN inference times include the full training pipeline (pre-processing, forward pass, and post-processing to obtain feasible solutions). The measurements reflect the time required to process a batch of $256$ samples, averaged for $10,000$ runs. From the table, the DNN-based precoding for all parameterizations is significantly faster than the WMMSE algorithm, which relies on many iterations to converge, and in each iteration the algorithm applies complex operations including matrix inversions. 

Among the DL models, $\mathcal P_{ri}$ exhibits the lowest latency, followed by $\mathcal P_{ncv}$ then $\mathcal P_{cps}$, while $\mathcal P_{hsc}$ is the most complex model. Interestingly, while $\mathcal{P}_{ncv}$ has a higher parameter count than $\mathcal{P}_{cps}$, it requires a slightly lower E2E inference time. This is attributed to the additional processing overhead in $\mathcal{P}_{cps}$ required for global phase removal. However, this latency penalty is marginal and is well-justified by the substantial performance and generalization gains offered by the $\mathcal P_{cps}$ parameterization. 
\section{Concluding Remarks}\label{sec:conclusions}
We established a DL framework for MU-MISO precoding design using CPS parameterizations. The proposed approach enables learning directly on the manifold of geometry-aware and physically distinct channel-precoder pairs, thereby overcoming phase-related ambiguities inherent in traditional representations. As a result, the proposed framework achieves higher sum rates and improved generalization with negligible additional complexity. Moreover, for the DNN considered, real-valued $\mathbb{CP}^n$ embeddings outperform hyperspherical coordinates, underscoring the importance of geometry-aware yet optimization-friendly parameterizations.

\begin{table}[]
\caption{E2E Latency Measurements.}
\label{tab:inference}
\centering
\begin{tabular}{|c|c|c|c|c|c|}
\hline
Precoding scheme & WMMSE  & $\mathcal P_{ri}$ & $\mathcal P_{ncv}$ & $\mathcal P_{cps}$ & $\mathcal P_{hsc}$ \\ \hline
CPU time (msec)  & $133.53$ & $2.87$              & $3.56$                & $3.77$               & $5.77$               \\ \hline
GPU time (msec)  & -      & $2.01$              & $2.42$                & $2.70$               & $4.15$               \\ \hline
\end{tabular}
\end{table}
\section*{Appendix}
We proceed by induction starting from the end of the vector. For the base case $m=n-1$, applying the definitions in (\ref{eq:hsc}) on $\left\|\mathbf u_{m+1:}\right\|$ results in:
\begin{align}
    \left\| \mathbf u_{n:}\right\| = & \sqrt{\cos^2{\theta_n} \prod_{i=1}^{n-1} \sin^2{\theta_i} + \sin^2{\theta_n}\prod_{i=1}^{n-1} \sin^2{\theta_i}} \nonumber \\ = & \prod_{i=1}^{n-1} \sin{\theta_i}.
\end{align}
Extending this via backward induction, one can express the tail norm of $\mathbf u$ for any index $m$ as:
\begin{equation}\label{eq:tail}
    \left\|\mathbf u_{m+1:} \right\| = \prod_{i=1}^{m} \sin{\theta_i}.
\end{equation}
Substituting the result in (\ref{eq:tail}) and the expression of $|u_m|$ into the right hand side of (\ref{eq:theta}) yields the following:
\begin{align}
    \arctan{\left(\frac{\left\|\mathbf u_{m+1:}\right\|}{|u_m|}\right)} = \arctan{\left(\frac{\sin{\theta_m} \prod_{i=1}^{m-1} \sin{\theta_i}}{\cos{\theta_m}\prod_{i=1}^{m-1} \sin{\theta_i}}\right)} = \theta_m.
\end{align}
This completes the proof.

\bibliographystyle{IEEEtran}
\bibliography{references}	

@article{shi2011iteratively,
  title={An iteratively weighted {MMSE} approach to distributed sum-utility maximization for a {MIMO} interfering broadcast channel},
  author={Shi, Qingjiang and others},
  journal={IEEE Trans. Signal Process.},
  volume={59},
  number={9},
  pages={4331--4340},
  year={Sep. 2011},
  publisher={IEEE}
}

@article{zhang2022deep,
  title={A deep learning-based framework for low complexity multiuser {MIMO} precoding design},
  author={Zhang, Maojun and Gao, Jiabao and Zhong, Caijun},
  journal={IEEE Trans. Wireless Commun.},
  volume={21},
  number={12},
  pages={11193--11206},
  year={2022},
  publisher={IEEE}
}

@article{shi2023robust,
  title={Robust {WMMSE} precoder with deep learning design for massive {MIMO}},
  author={Shi, Junchao and Lu, An-An and Zhong, Wen and Gao, Xiqi and Li, Geoffrey Ye},
  journal={IEEE Trans. Commun.},
  volume={71},
  number={7},
  pages={3963--3976},
  year={2023},
  publisher={IEEE}
}

@article{kim2020deep,
  title={Deep learning methods for universal {MISO} beamforming},
  author={Kim, Junbeom and Lee, Hoon and Hong, Seung-Eun and Park, Seok-Hwan},
  journal={IEEE Wireless Commun. Lett.},
  volume={9},
  number={11},
  pages={1894--1898},
  year={2020},
  publisher={IEEE}
}

@article{abdullah2025impact,
  author    = {Abdullah, Zaid and others},
  title     = {The Impact of Channel Representation on Deep Learning-Based {MIMO} Precoding},
  journal   = {TechRxiv},
  year      = {2025},
  month     = {Dec.},
  doi       = {10.36227/techrxiv.176472696.64455714/v1},
  note      = {Available at \url{https://www.techrxiv.org/1364921}}
}

@article{shi2021deep,
  title={Deep learning-based robust precoding for massive {MIMO}},
  author={Shi, Junchao and Wang, Wenjin and Yi, Xinping and Gao, Xiqi and Li, Geoffrey Ye},
  journal={IEEE Trans. Commun.},
  volume={69},
  number={11},
  pages={7429--7443},
  year={2021},
  publisher={IEEE}
}

@article{vu2021machine,
  title={Machine learning-enabled joint antenna selection and precoding design: From offline complexity to online performance},
  author={Vu, Thang X and others},
  journal={IEEE Trans. Wireless Commun.},
  volume={20},
  number={6},
  pages={3710--3722},
  year={2021},
  publisher={IEEE}
}

@article{wang2025learning,
  title={Learning-based joint antenna selection and precoding design for cell-free {MIMO} networks},
  author={Wang, Liangzhi and Chen, Chen and Zhang, Jie and Fischione, Carlo},
  journal={IEEE Trans. Veh. Technol.},
  year={2025},
  publisher={IEEE}
}

@article{liu2025deep,
  title={Deep Learning-Based Low-Complexity Hybrid Precoding for Massive {MIMO} Systems},
  author={Liu, Qingli and Wu, Jiayu and Cao, Yewen},
  journal={IEEE Commun. Lett.},
  volume={29},
  number={10},
  month={Oct.},
  year={2025},
  pages={2238--2242},
  publisher={IEEE}
}

@article{elbir2019hybrid,
  title={Hybrid precoding for multiuser millimeter wave massive {MIMO} systems: {A} deep learning approach},
  author={Elbir, Ahmet M and Papazafeiropoulos, Anastasios K},
  journal={IEEE Trans. Veh. Technol.},
  volume={69},
  number={1},
  pages={552--563},
  month = {Jan.},
  year={2020},
  publisher={IEEE}
}

@article{liu2022dl,
  title={{DL}-based energy-efficient hybrid precoding for mmwave massive {MIMO} systems},
  author={Liu, Fulai and Zhang, Lijie and Yang, Xianghuan and Li, Tiangui and Du, Ruiyan},
  journal={IEEE Trans. Veh. Technol.},
  volume={72},
  number={5},
  pages={6103--6112},
  month = {May},
  year={2023},
  publisher={IEEE}
}

@book{bengtsson2017geometry,
  title={Geometry of quantum states: an introduction to quantum entanglement},
  author={Bengtsson, Ingemar and {\.Z}yczkowski, Karol},
  year={2017},
  publisher={Cambridge Univ. Press}
}

@article{kingma2014adam,
  title={Adam: A method for stochastic optimization},
  author={Kingma, Diederik P and Ba, Jimmy L.},
  journal={arXiv preprint},
  year={2014},
  url = "https://arxiv.org/pdf/1412.6980"
}

@InProceedings{SB-9931,
    author = {Mats Bengtsson and Bj{\"o}rn Ottersten},
    title = {Optimal Downlink Beamforming Using Semidefinite Optimization},
    booktitle = {Proc. 37th Annual Allerton Conference on Communication, Control, and Computing},
    pages = {987-996},
    year = {1999},
    month = Sep,
    note = {Invited paper}
    }

@InCollection{SB-0105,
    author = {Mats Bengtsson and Bj{\"o}rn Ottersten},
    title = {Optimal and Suboptimal Transmit Beamforming},
    booktitle = {Handbook of Antennas in Wireless Communications},
    year = {2001},
    month = Aug,
    editor = {Lal C. Godara},
    publisher = {CRC Press}
    }

@ARTICLE{6832894,
    author={Bjornson, E. and Bengtsson, M. and Ottersten, B.},
    journal={Signal Processing Magazine, IEEE},
    title={Optimal Multiuser Transmit Beamforming: A Difficult Problem with a Simple Solution Structure [Lecture Notes]},
    year={2014},
    month={July},
    volume={31},
    number={4},
    pages={142-148},
    doi={10.1109/MSP.2014.2312183},
    ISSN={1053-5888}
}

@article{abdullah2022low,
  title={Low-complexity antenna selection and discrete phase-shifts design in {IRS}-assisted multiuser massive {MIMO} networks},
  author={Abdullah, Zaid and Chen, Gaojie and Lambotharan, Sangarapillai and Chambers, Jonathon A},
  journal={IEEE Trans. Veh. Technol.},
  volume={71},
  number={4},
  pages={3980--3994},
  year={2022},
  publisher={IEEE}
}

@article{yang2013performance,
  title={Performance of conjugate and zero-forcing beamforming in large-scale antenna systems},
  author={Yang, Hong and Marzetta, Thomas L},
  journal={IEEE J. Sel. Areas Commun.},
  volume={31},
  number={2},
  pages={172--179},
  year={2013},
  publisher={IEEE}
}

@misc{github_mimo_cps,
  author       = {Zaid Abdullah},
  title        = {mimo\_cps},
  year         = {2026},
  publisher    = {GitHub},
  howpublished={GitHub repository,    \url{https://github.com/zaid-89/mimo_cps}},
  note         = {{A}ccessed 2026-03-06}
}
\end{document}